\documentclass{article}
\usepackage{tikz}
\usepackage{subcaption}
\usepackage[utf8]{inputenc}
\usepackage{booktabs}
\usepackage[accepted]{icml2019}
\usepackage{graphicx}
\usepackage{multirow}
\usepackage{amsmath}

\usepackage{hyperref}
\hypersetup{
    colorlinks=false,
    linkcolor=black,
    filecolor=black,      
    urlcolor=black,
}

\date{}

\begin{document}
    \twocolumn[
    \icmltitle{Neural Logic Reinforcement Learning}

    \begin{icmlauthorlist}
    \icmlauthor{Zhengyao Jiang}{liv}
    \icmlauthor{Shan Luo}{liv}
    \end{icmlauthorlist}

    \icmlaffiliation{liv}{Department of Computer Science, University of Liverpool, Liverpool, United Kingdom}
    
    \icmlcorrespondingauthor{Zhengyao Jiang}{z.jiang22@student.liverpool.ac.uk}
    \icmlcorrespondingauthor{Shan Luo}{shan.luo@liverpool.ac.uk}

    \vskip 0.3in
    ]


\printAffiliationsAndNotice{}  

    \begin{abstract}
        Deep reinforcement learning (DRL) has achieved significant breakthroughs in various tasks. However, most DRL algorithms suffer a problem of generalising the learned policy, which makes the policy performance largely affected even by minor modifications of the training environment. Except that, the use of deep neural networks makes the learned policies hard to be interpretable. To address these two challenges, we propose a novel algorithm named Neural Logic Reinforcement Learning (NLRL) to represent the policies in reinforcement learning by first-order logic. NLRL is based on policy gradient methods and differentiable inductive logic programming that have demonstrated significant advantages in terms of interpretability and generalisability in supervised tasks. Extensive experiments conducted on cliff-walking and blocks manipulation tasks demonstrate that NLRL can induce interpretable policies achieving near-optimal performance while showing good generalisability to environments of different initial states and problem sizes.
    \end{abstract}
    \section{Introduction}
In recent years, Deep Reinforcement Learning (DRL) algorithms have achieved stunning breakthroughs in vairous tasks, e.g., video game playing \cite{Mnih2015} and the game of Go \cite{Silver2017}. However, similar to traditional reinforcement learning algorithms such as tabular TD-learning \cite{Sutton1998}, DRL algorithms can only learn policies that are hard to interpret \cite{MONTAVON2018} and cannot be generalized from one environment to another similar one \cite{Wulfmeier2017}.

\cite{Finale2017} defines interpretability as the ability to explain or to present the decision in understandable terms.
The interpretability is a critical capability of reinforcement learning or generally all machine learning algorithms for system verification and improvement. Interpretable RL can promote the scientific understanding of the algorithm or the problem to be solved. Furthermore, we can check whether an AI system is safe and whether it complies with existing rules, ethnically or legally, in human society. On the engineering level, the interpretability also enables 
easier debugging of the system. The neural network based DRL models, however, lack interpretability since the inference processes of neural networks are opaque to humans.

The generalizability is also important for reinforcement learning algorithms. In the real world, it is not common that the training and test environments are exactly the same. However, most DRL algorithms have the assumption that these two environments are identical, which makes a trained network that performs well on one task often performs very poorly on a new but similar task.
An example
is the reality gap \cite{Collins2018} in the robotics applications that often makes agents trained in simulation 
ineffective once transferred in the real world.

The generalization of DRL policy is a rather intricate and difficult problem since the action can affect the environment dynamics.
In supervised learning, there are regularization techniques such as Dropout \cite{srivastava2014dropout} that can help to promote generalization. However, \cite{Zhang2018} shows that noise injection methods used in several DRL works cannot robustly detect or alleviate overfitting. On the other hand, it is good to see that, similar to the supervised learning, a proper inductive bias that fits the problem bias can significantly improve the generalizability of the learned policies \cite{Zhang2018}. One candidate of inductive bias suitable for general decision-making is the relational inductive bias \cite{Zambaldi2018}. Relational inductive bias usually represents the abstract concepts as entities and relationships between them and perform deduction on the relations. The graph-based relational inductive bias has been tested in RL context \cite{Zambaldi2018,Wang2018} and showed significantly better generalization compared with Multilayer Perceptron (MLP) architectures or Convoluational Neural Networks (CNNs). However, another more expressive relational inductive, i.e., the one based on first-order logic, is still not much explored by the DRL community.

The traditional symbolic methods intrinsically have good interpretable and generalizable capabilities, however, require the systems dynamics to be known and ideally deterministic when solving general sequential decision-making problems~\cite{Fikes1971}.
By contrast, relational reinforcement learning \cite{Dzeroski2001} learns first-order logic rules in some simple block manipulation tasks without the knowledge of system dynamics, and also shows good generalizability and interpretability on these tasks.
However, such methods become ineffective when applied to more complex tasks, especially given that the symbolic learning methods have poor scalability.

A spectrum of such interpretable neural architectures is Differentiable Inductive Logic Programming (DILP)~\cite{Rock2017,cohen2017tensorlog,Evans2018}. Compared with traditional symbolic logic induction methods, with the use of differentiable models, DILP can leverage modern gradients based methods.
On the other side, thanks to the strong relational inductive bias, DILP shows superior interpretability and generalization ability than neural networks~\cite{Evans2018}. However, to the authors' best knowledge, all current DILP algorithms are only tested in supervised tasks such as hand-crafted concept learning~\cite{Evans2018} and knowledge base completion~\cite{Rock2017,cohen2017tensorlog}.

To make a step further, in this paper we develop a novel framework named as Neural Logic Reinforcement Learning (NLRL) to enable the differentiable induction in sequential decision-making tasks. It can alleviate the interpretability and generalizability problems in deep reinforcement learning. In addition, the proposed NLRL framework is also of great significance in advancing the DILP research. By applying DILP in sequential decision-making tasks, the agents can learn new concepts without human supervision, instead of describing a concept already known to the human in supervised learning tasks. 


The rest of the paper is organized as follows: In Section \ref{sec:related}, related works are reviewed and discussed; In Section \ref{sec:preliminary}, an introduction to the preliminary knowledge is presented, including the first-order logic programming $\partial$ILP and Markov Decision Processes (MDPs); In Section \ref{sec:model}, the NLRL model is introduced, both the DILP architecture and a general NLRL framework modeled with MDPs; In Section \ref{sec:exp}, the experiments of NLRL on block manipulation and cliff-walking are presented; In the last section, the paper is concluded and future directions are directed.
    \section{Related Work} \label{sec:related}
We place our work in the development of relational reinforcement learning \cite{Dzeroski2001} that represent states, actions and policies in Markov Decision Processes (MDPs) using the first order logic where transitions and rewards structures of MDPs are unknown to the agent. To this end, in this section we review the evolution of relational reinforcement learning and highlight the differences of our proposed NLRL framework with other algorithms in relational reinforcement learning.

Early attempts that represent states by first-order logics in MDPs appeared at the beginning of this century \cite{Boutilier2001,Yoon2002,Guestrin2003}, however, these works focused on the situation that transitions and reward structures are known to the agent. In such cases with environment models known, variations of traditional MDP solvers such as dynamic programming \cite{Boutilier2001}, linear programming \cite{Guestrin2003} and heuristic greedy searching \cite{Yoon2002} were employed to optimise policies in training tasks that can be generalized to large problems. In these works, the transition and reward functions are also represented in logic forms. The setting limits their application to complex tasks whose transition and reward functions are hard to be modeled using the first order logic.


The concept of relational reinforcement learning was first proposed by \cite{Dzeroski2001} in which the first order logic was first used in reinforcement learning. There are extensions of this work \cite{Driessens2003,Driessens2004}, however, all these algorithms employ non-differential operations, which makes it hard to apply new breakthroughs happened in DRL community. In contrast, in our work using differentiable inductive logic programming, once given the logic interpretations of states and actions, any type of MDPs can be solved with policy gradient methods compatible with DRL algorithms. Furthermore, most relational reinforcement learning algorithms represent the induced policy in a single clause and some auxiliary predicates, e.g., the predicates that count the number of blocks, are given to the agent. In our work, the DILP algorithms have the ability to learn the auxiliary invented predicates by the agents themselves, which not only enables stronger expressive ability but also gives possibilities for knowledge transfer.



One previous work close to ours is \cite{gretton2007} that also trains the parameterised rule-based policy using policy gradient. An approach was proposed to pre-construct a set of potential policies in a brutal force manner and train the weights assigned to them using policy gradient. Compared to this work, in our NLRL framework weights are not assigned directly to the whole policy; the parameters to be trained are involved in the deduction process and the number of parameters is significantly smaller than that in an enumeration of all policies, especially for larger problems, which gives our method better scalability. In addition, in \cite{gretton2007}, expert domain knowledge is needed to specify the potential rules for the exact task that the agent is dealing with. However, in our work, we use the same rules templates for all tasks we test on, which means all the potential rules have the same format across tasks.

A recent work on the topic \cite{Zambaldi2018} proposes deep reinforcement learning with relational inductive bias that applies neural network mixed with self-attention to reinforcement learning tasks and achieves the state-of-the-art performance on the StarCraftII mini-games. The proposed methods show some level of generalization ability on the constructed block world problems and StarCraft mini-games, showing the potential of relation inductive bias in larger problems. However, as a graph-based relational model was used \cite{Zambaldi2018}, the learned policy is not fully explainable and the rules expression is limited, different from the interpretable logic-represented policies learned in ours using DILP. 

A parallel work~\cite{dong2019} use neural networks to approximate the first-order logic deduction, achieving good generalization on both supervised and reinforcement learning tasks. Their method has better scalibility than DILP based approaches, however, it is still not clear how to interpret the policies learned by their model.
    \section{Preliminary} \label{sec:preliminary}
In this section, we give a brief introduction to the necessary background knowledge of the proposed NLRL framework.
Basic concepts of the first-order logic are first introduced. $\partial$ILP, a DILP model that our work is based on, is then described. The Markov Decision Process (MDP) and reinforcement learning are also briefly introduced.

\subsection{First-Order Logic Programming}
Logic programming languages are a class of programming languages using logic rules rather than imperative commands.
One of the most famous logic programming languages is ProLog, which expresses rules using the first-order logic.
In this paper, we use the subset of ProLog, i.e., DataLog \cite{Getoor2007}. 

\textbf{Predicate names} (or for short, predicates), \textbf{constants} and \textbf{variables} are three primitives in DataLog. In the language of relational learning, a predicate name is also called a relation name, and a constant is also termed as an entity \cite{Getoor2007}.
An \textbf{atom} $\alpha$ is a predicate followed by a tuple $p(t_1, ..., t_n)$, where $p$ is an n-ary predicate and $t_1, ..., t_n$ are \textbf{terms}, either variables or constants.
For example, in the atom \textit{father(cart, Y)}, \textit{father} is the predicate name, \textit{cart} is a constant and \textit{Y} is a variable.
If all terms in an atom are constants, this atom is called a \textbf{ground atom}.
We denote the set of all ground atoms as $G$.
A predicate can be defined by a set of ground atoms, in which case the predicate is called an \textbf{extensional predicate}.
Another way to define a predicate is to use a set of \textbf{clauses}.
A clause is a rule in the form $\alpha \leftarrow \alpha_1, ..., \alpha_n$, where $\alpha$ is the \textbf{head} atom and $\alpha_1,..., \alpha_n$ are body atoms.
The predicates defined by clauses are termed as \textbf{intensional predicates}.

\subsection{$\partial$ILP}
Inductive logic programming (ILP) is a task to find a definition (set of clauses) of some intensional predicates, given some positive examples and negative examples \cite{Getoor2007}.
The attempts that combine ILP with differentiable programming are presented in \cite{Evans2018,Rock2017} and $\partial$ILP \cite{Evans2018} is introduced here that our work is based on.

The major component of $\partial$ILP operates on the valuation vectors $\boldsymbol{e}$ whose space is $E=[0,1]^{|G|}$, where each element of a valuation vector represents the confidence that a related ground atom is true. The logical deduction of each step of the $\partial$ILP is applied to the valuation vector. The new facts are derived from the facts provided by the valuation vector in the last step. For each predicate, $\partial$ILP generates a series of potential clauses combinations in advance based on rules templates. Trainable weights are assigned to clauses combinations, and the sum of weights for a predicate is constrained to be summed up to 1 using a softmax function.

With the differentiable deduction, the system can be trained with gradient-based methods. The loss value is defined as the cross-entropy between the confidence of predicted atoms and the ground truth. 
Compared with traditional inductive logic programming methods, $\partial$ILP has advantages in terms of robustness against noise and ability to deal with fuzzy data \cite{Evans2018}.

\ifx
\subsection{Markov Decision Processes}
The Markov Decision Process (MDP) \cite{Puterman1994} can be described as a tuple $(S, A, T, R)$, where, $S$ is a set of states; $A$ is a set of actions; $T(s,a,s')$ is the transition function; $R(s,a)$ is the reward function.

An MDP starts at a state $s_0$ and in each step $t$:
(1) the agent chooses an action $a_t \in A$,
and then gets an reward $r_t = R(s_t, a_t)$ and
(2) the environment is transformed to a new state $s_{t+1}$ following the probability $T(s_t, a_t, s_{t+1})$.
The outcome of an MDP is a trajectory $(s_1,a_1,s_2,a_2,...)$.
To solve an MDP is to find a policy $\pi$ that maps states to the distribution of actions, optimizing the expectation of the sum of rewards.

MDPs can be solved by dynamic programming methods such as value iteration \cite{Bellman1954} and policy iteration \cite{Howard1960,Puterman1994} when the transition model and the reward function are known to the agent.
When the reward function and/or exact transition probability are unknown, it is possible to use \emph{reinforcement learning (RL)} techniques, such as
temporal-difference learning \cite{Sutton1988}, policy gradient \cite{Willia1992} and actor-critic methods \cite{NIPS1999_1786}.
\fi
    \section{Neural Logic Reinforcement Learning} \label{sec:model}
In this section, the details of the proposed NLRL framework\footnote{Code available at the homepage of the paper: \url{https://github.com/ZhengyaoJiang/NLRL}} are presented. A new DILP architecture termed as Differentiable Recurrent Logic Machine (DRLM), an improved version of $\partial$ILP, is first introduced.
The MDP with logic interpretation is then proposed to train the DILP architecture.  

\subsection{Differentiable Recurrent Logic Machine}

Recall that $\partial$ILP operates on the valuation vectors whose space is $E=[0,1]^{|G|}$, each element of which represents the confidence that a related ground atom is true.
A DRLM is a mapping $f_{\boldsymbol{\theta}}: E \to E$, which performs the deduction of the facts $\boldsymbol{e}_0$ using weights $\boldsymbol{w}$ associated with possible clauses. $f_{\boldsymbol{\theta}}$ can then be decomposed into repeated application of single step deduction functions $g_{\boldsymbol{\theta}}$, namely,
\begin{equation}
f_{\boldsymbol{\theta}}^t(\boldsymbol{e}_0) =\begin{cases}
    g_{\boldsymbol{\theta}}(f_{\boldsymbol{\theta}}^{t-1}(\boldsymbol{e}_0)), & \text{if $t>0$}.\\
    \boldsymbol{e}_0, & \text{if $t=0$}.
  \end{cases}
\text{,}
\end{equation}
where $t$ is the deduction step. $g_{\boldsymbol{\theta}}$ implements one step deduction of all the possible clauses weighted by their confidences. We denote the probabilistic sum as $\oplus$ and
\begin{equation}
\boldsymbol{a} \oplus \boldsymbol{b} = \boldsymbol{a} + \boldsymbol{b} - \boldsymbol{a} \odot \boldsymbol{b}
\text{,}
\end{equation}
where $\boldsymbol{a} \in E, \boldsymbol{b} \in E$. $g_{\boldsymbol{\theta}}$ can then be expressed as
\begin{equation}
g_{\boldsymbol{\theta}}(\boldsymbol{e}) = \left( \sum_n^{\oplus} \sum_j w_{n,j} h_{n,j}(\boldsymbol{e}) \right)+\boldsymbol{e}_0
\text{,}
\label{eq:g}
\end{equation}
where $h_{n,j}(\boldsymbol{e})$ implements one-step deduction of the valuation vector $\boldsymbol{e}$ using $j$th possible definition of $n$th clause.\footnote{Computational optimization is to replace $\oplus$ with typical + when combining valuations of two different predicates. For further details on the computation of $h_{n,j}(\boldsymbol{e})$ ($F_c$ in the original paper), readers are referred to Section 4.5 in \cite{Evans2018}.}
For every single clause $c$, we can constrain the sum of its weights to be 1 by letting $\boldsymbol{w}_c = \text{softmax}(\boldsymbol{\theta}_c)$, where $\boldsymbol{w}_c$ is the vector of weights associated with the predicate $c$ and $\boldsymbol{\theta}_c$ are related parameters to be trained. 

Compared to $\partial$ILP, in DRLM the number of clauses used to define a predicate is more flexible thanks to associating the weights with clauses directly instead of combinations of clauses; it needs less memory to construct a model (less than 10 GB in all our experiments); it also enables learning longer logic chaining of different intensional predicates. All these benefits make the architecture be able to work in larger problems. Detailed discussions on the modifications and their effects can be found in the appendix.

\subsection{Markov Decision Process with Logic Interpretation}
In this section, we present a formulation of MDPs with logic interpretation and show how to solve the MDP with the combination of policy gradient and DILP.

An MDP with logic interpretation is a triple $(M, p_S, p_A)$:
\begin{itemize}
    \item $M=(S,A,T,R)$ is a finite-horizon MDP;
    \item $p_S: S \to 2^G$ is the state encoder that maps each state to a set of atoms including both information of the current state and background knowledge;
    \item $p_A: [0,1]^{|D|} \to [0,1]^{|A|}$ is the action decoder that maps the valuation (or score) of a set of atoms $D$ to the probability of actions.
\end{itemize}

For a DILP system $f_{\boldsymbol{\theta}}: 2^G \to [0,1]^{|D|}$, the policy $\pi: S \to [0,1]^{|D|}$ can be expressed as $\pi(s) = p_A(f_{\boldsymbol{\theta}}(p_S(s)))$.
Thus any policy-gradient methods applied to DRL can also work for DILP.
$p_S$ and $p_A$ can either be hand-crafted or represented by neural architectures.
The action selection mechanism in this work is to add a set of action predicates into the architecture, which depends on the valuation of these action atoms. Therefore, the action atoms should be a subset of $D$. As for $\partial$ILP, valuations of all the atoms will be deduced, i.e., $D=G$. 
If $p_S$ and $p_A$ are neural architectures, they can be trained together with the DILP architectures. $p_S$ extracts entities and their relations from the raw sensory data. In addition, the use of a neural network to represent $p_A$ enables agents to make decisions in a more flexible manner. For instance, the output actions can be deterministic and the final choice of action may depend on more atoms rather than only action atoms if the optimal policy cannot be easily expressed as first-order logic. For simplicity, in this work, we will only use the hand-crafted $p_S$ and $p_A$.
Notably, $p_A$ is required to be differentiable so that we can train the system with policy gradient methods operating on discrete, stochastic action spaces, such as vanilla policy gradient~\cite{Willia1992}, A3C~\cite{Mnih2016}, TRPO~\cite{Schulman2015} or PPO~\cite{Schulman2017ProximalPO}.

We use the following schema to represent the $p_A$ in all experiments. Let $p_A(a|\boldsymbol{e})$ be the probability of choosing action $a$ given the valuations $\boldsymbol{e} \in [0,1]^{|D|}$.
The probability of choosing an action $a$ is proportional to its valuation if the sum of the valuation of all action atoms is larger than 1; otherwise, the difference between 1 and the total valuation will be evenly distributed to all actions, i.e.,
\begin{equation}
    p_A(a|\boldsymbol{e})= 
\begin{cases}
    \frac{l(e,a)}{\sigma},    &  \sigma \geq 1\\
    l(e,a)+\frac{1-\sigma}{|A|},&  \sigma < 1
\end{cases}
\end{equation}
where $l: [0,1]^{|D|} \times A \to [0,1]$ maps from valuation vector and action to the valuation of that action atom, $\sigma$ is the sum of all action valuations $\sigma = \sum_{a} l(e,a)$.
Empirically, this design is crucial for inducing an interpretable and generalizable policy. If we apply a trivial normalization, it is not necessary for NLRL agent to increase rule weights to 1 for the sake of exploitation. 
The agent instead only needs to keep the relative valuation advantages of desired actions over other actions, which in practice leads to tricky policies.
We train all the agents with vanilla policy gradient \cite{Willia1992} in this work.
    \section{Experiments and Analysis} \label{sec:exp}
In general, the experiment is going to act as empirical investigations of the following hypothesis:
\begin{enumerate}
    \item NLRL can learn policies that are comparable to neural networks in terms of expected returns;
    \item To induce these policies, we only need to inject minimal background knowledge;
    \item The induced policies are explainable;
    \item The induced policies can generalize to environments that are different from the training environments in terms of scale or initial state. 
\end{enumerate}

Four sets of experiments, which are \textit{STACK}, \textit{UNSTACK} and \textit{ON} block manipulation tasks, and cliff-walking, have been conducted and the benchmark model is a fully-connected neural network. 
The induced policy will be evaluated in terms of expected returns, generalizability and interpretability.

\subsection{Experiment Setup}
In the experiments, to test the robustness of the proposed NLRL framework, we only provide minimal atoms describing the background and states while the auxiliary predicates 
are not provided. The agent must learn auxiliary invented predicates by themselves, together with the action predicates.

\subsubsection{Block Manipulation}
In this environment, the agent will learn how to stack the blocks into certain styles, that are widely used as a benchmark problem in the relational reinforcement learning research. We examine the performance of the agent on three subtasks: \textit{STACK}, \textit{UNSTACK} and \textit{ON}. In the \textit{STACK} task, the agent needs to stack the scattered blocks into a single column. In the \textit{UNSTACK} task, the agent needs to do the opposite operation, i.e., spread the blocks on the floor. In the \textit{ON} task, it is required to put a specific block onto another one. In all three tasks, the agent can only move the topmost block in a pile of blocks. When the agent finishes its goal it will get a reward of 1; before that, the agent keeps receiving a small penalty of -0.02. The training is terminated if the agent does not reach the goal within 50 steps.

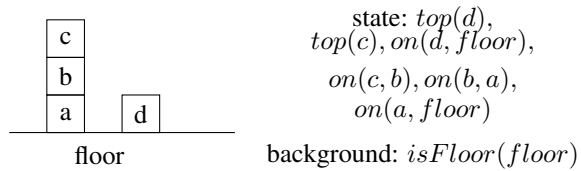
\begin{figure}[h]
    \begin{center}
    \begin{tikzpicture}
        \draw (0,0) -- (3,0);
        \draw (0.5,0) rectangle (1.0,0.5) node[pos=.5] {a};
        \draw (0.5,0.5) rectangle (1.0,1.0) node[pos=.5] {b};
        \draw (0.5,1.0) rectangle (1.0,1.5) node[pos=.5] {c};
        \draw (1.5,0) rectangle (2.0,0.5) node[pos=.5] {d};
        \node[](floor) at (1.2, -0.3) {floor};

        \node[](text) at (5.5, 1.5) {state: $top(d),$};
        \node[](text) at (5.5, 1.2) {$top(c),on(d,floor),$};
        \node[](text) at (5.5, 0.7) {$on(c,b),on(b,a),$};
        \node[](text) at (5.5, 0.3) {$on(a,floor)$};
        
        \node[](text2) at (5.5, -0.3) {background: $isFloor(floor)$};
    \end{tikzpicture}
    \caption{A Blocks Manipulation state noted as $((a,b,c),(d))$.} \label{fig:blocks}
    \end{center}
\end{figure}

There are 5 different entities, 4 blocks labeled as \textit{a, b, c, d} and \textit{floor}.
The state predicates are \textit{on(X,Y)} and \textit{top(X)}.
\textit{on(X,Y)} means the block \textit{X} is on the entity \textit{Y} (either blocks or floor).
$top(X)$ means the block $X$ is on top of an column of blocks.
Notably, $top(X)$ cannot be expressed using $on$ here as in DataLog there is no expression of negation, i.e., it cannot have ``$top(X)$ means there is no $on(Y,X)$ for all $Y$".
For all tasks, a common background knowledge is \textit{isFloor(floor)},
and for the \textit{ON} task, there is one more background knowledge predicate \textit{goalOn(a,b)}, which indicates the target is to move block $a$ onto the block $b$.
The action predicate is \textit{move(X,Y)} and there are 25 actions atoms in this task.
The action is valid only if both $Y$ and $X$ are on the top of a pile or $Y$ is $floor$ and $X$ is on the top of a pile. If the agent chooses an invalid action, e.g., \textit{move(floor, a)}, the action will not make any change to the state. We use a tuple of tuples to represent the states, 
where each inner tuple represents a column of blocks, from bottom to top. For instance, Figure~\ref{fig:blocks} shows the state $((a,b,c),(d))$ and its logic representation.

The training environment of the \textit{UNSTACK} task starts from a single column of blocks $((a, b, c, d))$. To test the generalizability of the induced policy, we construct the test environment by modifying its initial state by swapping the top 2 blocks or dividing the blocks into 2 columns. The agent is also tested in the environments with more blocks stacking in one column.
Therefore, the initial states of all the generalization test of \textit{UNSTACK} are:
$((a, b, d, c))$, $((a, b), (c, d))$, $((a, b, c, d, e))$, $((a, b, c, d, e, f))$ and $((a, b, c, d, e, f, g))$.
For the \textit{STACK} task, the initial state is $((a), (b), (c), (d))$ in training environment.
Similar to the \textit{UNSTACK} task, we swap the right two blocks, divide them into 2 columns and increase the number of blocks as generalization tests.
The initial states of all the generalization test of \textit{STACK} are: $((a), (b), (d), (c))$, $((a,b), (d,c))$, $((a), (b), (c), (d), (e))$, $((a), (b), (c), (d), (e), (f))$, $((a), (b), (c), (d), (e), (f), (g))$.
For \textit{ON}, the initial state is $((a, b, c, d))$. We swap either the top two or middle two blocks in this case, and also increase the total number of blocks.
The initial states of all the generalization test of \textit{ON} are thus: $((a, b, d, c))$, $((a, c, b, d))$, $((a, b, c, d, e))$, $((a, b, c, d, e, f))$ and $((a, b, c, d, e, f, g))$.

\subsubsection{Cliff-walking}

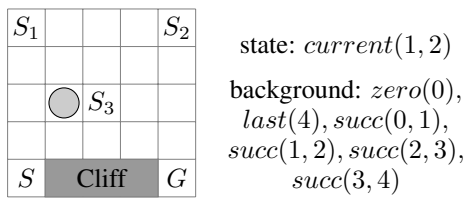
\begin{figure}[h]
    \begin{center}
    \begin{tikzpicture}
        \draw[step=0.5cm,gray,very thin] (0,0) grid (2.5,2.5);
        \filldraw[fill=black!40!white, draw=gray] (0.5,0) rectangle (2,0.5);
        \filldraw[fill=black!20!white, draw=black] (0.75, 1.25) circle (0.2);
        \node[](cliff) at (1.25,0.25){Cliff};
        \node[](S) at (0.25,0.25){$S$};
        \node[]() at (0.25,2.25){$S_1$};
        \node[]() at (2.25,2.25){$S_2$};
        \node[]() at (1.25,1.25){$S_3$};
        \node[](G) at (2.25,0.25){$G$};
        \node[](text) at (4.5, 2.0) {state: $current(1,2)$};
        \node[](text2) at (4.5, 1.4) {background: $zero(0),$};
        \node[](text2) at (4.5, 1.0) {$last(4), succ(0,1),$};
        \node[](text2) at (4.5, 0.6) {$succ(1,2), succ(2,3),$};
        \node[](text2) at (4.5, 0.2) {$succ(3,4)$};
    \end{tikzpicture}
    \caption{Cliff-walking, the circle represents location of the agent.} \label{fig:cliff}
    \end{center}
\end{figure}

Cliff-walking is a commonly used toy task for reinforcement learning.
We modify the version in \cite{Sutton1998} to a 5 by 5 field, as shown in Figure \ref{fig:cliff}. When the agent reaches the cliff position it gets a reward of -1, and if the agent arrives at the goal position, it gets a reward of 1. Before reaching these absorbing positions, the agent keeps receiving a small penalty of -0.02 at each step, encouraged to reach the goal as soon as possible. If the agent fails to reach the absorbing states within 50 steps, the game will be terminated. This problem can be modelled as a finite-horizon MDP.

The constants in this experiment are integers from 0 to 4.
We inject basic knowledge about natural numbers including the smallest number (\textit{zero(0)}), largest number (\textit{last(4)}), and the order of the numbers (\textit{succ(0,1), succ(1,2), ...}).
The symbolic representation of the state is \textit{current(X,Y)}, which specifies the current position of the agent.
There are four action atoms: \textit{up(), down(), left() and right()}.

In the training environment of cliff-walking, the agent starts from the bottom left corner, labelled as $S$ in Figure~\ref{fig:cliff}. In the generalization tests, we move the initial position to the top left, top right, and the centre of the field, labelled as $S_1, S_2$ and $S_3$ respectively.
Then we increase the size of the whole field to 6 by 6 and 7 by 7 without retraining.

We also test a stochastic variant of cliff-walking, i.e., windy cliff-walking, where the agent has a 10\% chance to move downwards no matter which action it takes. 

\subsubsection{Hyperparameters}
Similar to $\partial$ILP, we use RMSProp to train the agent, whose learning rate is set as 0.001. The generalized advantages ($\lambda=0.95$) \cite{Schulman2015-2} are applied to the value network where the value is estimated by a neural network with one 20-units hidden layer.

Like the architecture design to the neural network, the rules templates are important hyperparameters to the DILP algorithms. The rules template of a clause indicates the arity of the predicate (can be 0, 1, or 2) and the number of existential variables (usually pick from $\{0, 1, 2\}$).
It also specifies whether the body of a clause can contain other invented predicates. We represent a rule template as a tuple of its three parameters, such as $(2, 1, True)$, for the simplicity of expression. The rules templates of the DRLM are quite general and the optimal setting can be searched automatically. In this work, however, we use the same rules templates for invented predicates across all the tasks, each with only 1 clause, i.e., $(1,1,True), (1,2,False), (2,1,True), (2,1,True)$. 
The templates of action predicates vary in different tasks but it is easy to find a good one by exhaustive search. To this end, little domain knowledge is needed. For the \textit{UNSTACK} and \textit{STACK} tasks, the action predicate template is $(2, 1, True)$. For the \textit{ON} task, the action predicate templates are $(2, 1, True)$ and $(2, 0, True)$.
There are four action predicates in the cliff-walking task, we give all these predicates the same template $(3, 1, True)$.

\subsubsection{Benchmark Neural Network Agent}
In all the tasks, in addition to a random agent, we use an MLP agent as another benchmark that has two hidden layers with 20 units and 10 units respectively.
All the units in the hidden layers use a ReLU  \cite{Nair2010} activation function. For the cliff-walking task, the input is the coordinates of the current position of the agent. For block stacking tasks, the input is a $7 \times 7 \times 7$ tensor $X$, where $X_{x,y,i}=1$ if the block indexed as $i$ is in position $x, y$. We set each dimension of the tensor as 7 that is the maximum number of blocks used in the generalization test.

\subsection{Results and Analysis}
The performance of policy deduced by NLRL is stable against different random seeds once all the hyper-parameters are fixed, therefore, we only present the evaluation results of the policy trained in the first run for NLRL here.
For the neural network agent, we pick the agent that performs best in the training environment out of 5 runs.
The induced policy is also evaluated in terms of interpretability.

\begin{figure*}
\centering
\begin{subfigure}{.33\linewidth}
  \centering
  \includegraphics[width=\linewidth]{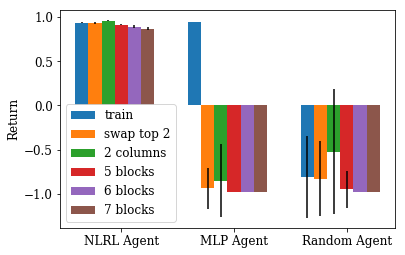}
  \caption{\textit{UNSTACK}}
\end{subfigure}
\begin{subfigure}{.33\linewidth}
  \centering
  \includegraphics[width=\linewidth]{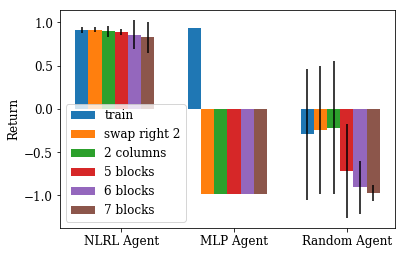}
  \caption{\textit{STACK}}
\end{subfigure}
\begin{subfigure}{.33\linewidth}
  \centering
  \includegraphics[width=\linewidth]{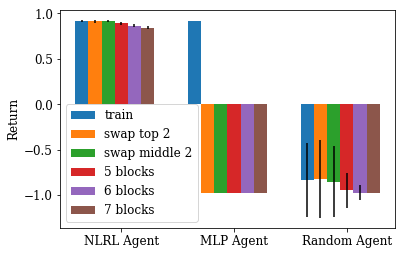}
  \caption{\textit{ON}}
\end{subfigure}
\begin{subfigure}{.33\linewidth}
  \centering
  \includegraphics[width=\linewidth]{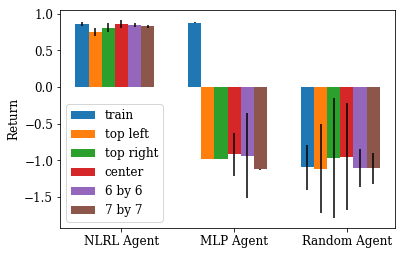}
  \caption{Cliff-walking}
\end{subfigure}
\begin{subfigure}{.33\linewidth}
  \centering
  \includegraphics[width=\linewidth]{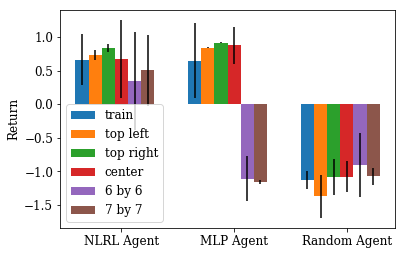}
  \caption{Windy Cliff-walking}
\end{subfigure}
\caption{Performance of different agents in the training and test environments. Each sub-figure shows the performance of the three agents in a task. The performance of each agent is divided into a group. In each group, the blue bar shows the performance in the training environment while the others show the performance in the test environments.}
\label{fig:performance}
\end{figure*}

\begin{table*}
  \centering
  \caption{The returns of different agents in the training and test environments. The first three columns demonstrate the returns of the three agents; the last column shows the returns of the optimal policy, which is computed using value iteration.}
  \small
\begin{tabular}{llllll}
\toprule
              &           &         NLRL &            MLP &        Random  & Optimal\\
\midrule
UNSTACK & training &  $0.937\pm0.008$ &      $\mathbf{0.940}\pm0.000$ &  $-0.807\pm0.466$ & $0.940$\\
              & swap top 2 &  $\mathbf{0.936}\pm0.009$ &   $-0.940\pm0.232$ &  $-0.827\pm0.428$ & $0.940$ \\
              & 2 columns &  $\mathbf{0.958}\pm0.006$ &  $-0.852\pm0.414$ &   $-0.522\pm0.710$ & $0.960$ \\
              & 5 blocks &   $\mathbf{0.915}\pm0.010$ &     $-0.980\pm0.000$ &  $-0.948\pm0.208$ & $0.920$\\
              & 6 blocks &  $\mathbf{0.891}\pm0.014$ &     $-0.980\pm0.000$ &     $-0.980\pm0.000$ & $0.900$\\
              & 7 blocks &  $\mathbf{0.868}\pm0.016$ &     $-0.980\pm0.000$ &     $-0.980\pm0.000$ & $0.880$\\
              \hline
STACK & training &   $0.910\pm0.033$ &      $\mathbf{0.940}\pm0.000$ &  $-0.292\pm0.759$ & $0.940$\\
              & swap right 2 &  $\mathbf{0.913}\pm0.029$ &     $-0.980\pm0.000$ &   $-0.240\pm0.739$ & $0.940$ \\
              & 2 columns &  $\mathbf{0.897}\pm0.064$ &     $-0.980\pm0.000$ &  $-0.215\pm0.772$ & $0.940$\\
              & 5 blocks &  $\mathbf{0.891}\pm0.032$ &     $-0.980\pm0.000$ &  $-0.718\pm0.542$ & $0.920$\\
              & 6 blocks &  $\mathbf{0.856}\pm0.169$ &     $-0.980\pm0.000$ &  $-0.905\pm0.307$ & $0.900$ \\
              & 7 blocks &  $\mathbf{0.828}\pm0.179$ &     $-0.980\pm0.000$ &  $-0.973\pm0.097$ & $0.880$\\
              \hline
ON & training &   $0.915\pm0.01$ &      $\mathbf{0.920}\pm0.000$ &  $-0.837\pm0.405$ & $0.920$\\
              & swap top 2 &  $\mathbf{0.912}\pm0.013$ &     $-0.980\pm0.000$ &  $-0.821\pm0.432$ & $0.920$\\
              & swap middle 2 &  $\mathbf{0.914}\pm0.011$ &     $-0.980\pm0.000$ &  $-0.853\pm0.394$ & $0.920$\\
              & 5 blocks &   $\mathbf{0.890}\pm0.016$ &     $-0.980\pm0.000$ &  $-0.949\pm0.195$ & $0.900$\\
              & 6 blocks &  $\mathbf{0.865}\pm0.018$ &     $-0.980\pm0.000$ &  $-0.975\pm0.081$ & $0.880$\\
              & 7 blocks &  $\mathbf{0.844}\pm0.017$ &     $-0.980\pm0.000$ &     $-0.980\pm0.000$ & $0.860$\\
              \hline
Cliff-walking & training &  $0.862\pm0.026$ &   $\mathbf{0.877}\pm0.008$ &  $-1.096\pm0.307$ & $0.880$\\
              & top left &  $\mathbf{0.749}\pm0.057$ &     $-0.980\pm0.000$ &  $-1.115\pm0.606$ & $0.840$\\
              & top right &  $\mathbf{0.809}\pm0.064$ &     $-0.980\pm0.000$ &  $-0.966\pm0.817$ & $0.920$\\
              & center &   $\mathbf{0.859}\pm0.05$ &  $-0.917\pm0.296$ &   $-0.952\pm0.730$ &$0.920$\\
              & 6 by 6 &  $\mathbf{0.841}\pm0.024$ &  $-0.934\pm0.578$ &   $-1.101\pm0.260$ & $0.860$\\
              & 7 by 7 &  $\mathbf{0.824}\pm0.024$ &  $-1.122\pm0.006$ &  $-1.107\pm0.209$ & $0.840$\\
              \hline
Windy Cliff-walking & training &  $\mathbf{0.663}\pm0.377$ &   $0.649\pm0.558$ &  $-1.129\pm0.135$ & $0.769\pm0.162$\\
              & top left &  $0.726\pm0.075$ &   $\mathbf{0.836}\pm0.008$ &   $-1.376\pm0.320$ & $0.837\pm0.068$\\
              & top right &  $0.834\pm0.061$ &   $\mathbf{0.919}\pm0.004$ &  $-1.089\pm0.266$ & $0.920\pm0.000$\\
              & center &  $0.672\pm0.579$ &   $\mathbf{0.859}\pm0.277$ &   $-1.082\pm0.230$ & $0.868\pm0.303$\\
              & 6 by 6 &  $\mathbf{0.345}\pm0.736$ &   $-1.110\pm0.335$ &  $-0.907\pm0.478$ & $0.748\pm0.135$\\
              & 7 by 7 &  $\mathbf{0.506}\pm0.528$ &  $-1.161\pm0.036$ &  $-1.077\pm0.129$ & $0.716\pm0.181$\\

\bottomrule
\end{tabular}
\label{tab}
\end{table*}

\subsubsection{Performance and Generalization Test}
The NLRL agent succeeds to find near-optimal policies on all the tasks. For generalization tests, we apply the learned policies on similar tasks, either with different initial states or problem sizes. 

We present the average and standard deviation of 500 repeats of evaluations in different environments in Figure~\ref{fig:performance} and Table~\ref{tab}. The highest average return of the three agents are marked in bold in each row of the table and the optimal performance of each task is also given. Each left group of bars in Figure~\ref{fig:performance} shows that the NLRL not only achieves a near-optimal performance in all the training environments but also successfully adapts to all the new environments we designed in experiments. In most of the generalization tests, the agents manage to keep the performance in the near optimal level even if they never experience these new environments before.
For instance, we can observe in Table~\ref{tab} that in the \textit{UNSTACK} task the NLRL agent achieves 0.937 average return, close to the optimal policy, and can achieve 0.940 final return. The minor difference between induced policy and the optimal one is caused by the stochasticity of the induced rules since the rule confidence is close but not exactly 1, which will be seen in the rules interpretations. When the top 2 blocks are swapped, the performance of NLRL agent is not affected. When the initial blocks are divided into 2 columns, it can still achieve 0.958 average return, very close to the optimal performance (0.960). The increase in the number of blocks gradually brings a larger difference between the return of the induced policy and the optimal one, whereas the difference is still less than 0.02.

The neural network agents learn optimal policy in the training environment of 3 block manipulation tasks and learn near-optimal policy in cliff-walking.
However, the neural network agent appears to only remember the best routes in the training environment rather than learn the general approaches to solving the problems. The overwhelming trend is, in varied environments, the neural networks perform even worse than a random player.

\subsubsection{Interpretation of the policies}
In all of five substasks we tested, the NLRL agent can induce human readable rules. We present the induced rules for two of them here, and these for other tasks can be found in the appendix.

\textbf{Induced policies for \textit{STACK}}: The policies induced by the NLRL agent in the \textit{STACK} task are:
\begin{equation}
    \begin{split}
        &0.964: pred1(X,Y) \leftarrow on(X,Z),top(Y) \\
        &0.970: pred2(X) \leftarrow on(X,Y),isFloor(Y) \\
        &0.923: pred4(X,Y) \leftarrow pred2(X),pred1(Y,X) \\
        &0.960: pred3(X) \leftarrow on(X,Y),pred1(Y,X) \\
        &0.903: move(X,Y) \leftarrow pred3(Y),pred4(X,Y)
    \end{split}
\end{equation}

In the learned policies, the agent uses several invented predicates, labelled as $pred1$, $pred2$, $pred3$ and $pred4$, to represent auxiliary concepts regarding the property of blocks.
The clause of $move$ is then constructed based on these concepts. 
The learned policies are interpreted in the forward chaining manner: the low level invented predicates are first interpreted whose body only contains existential predicates, following the predicates based on lower level predicates, and eventually the final clause of $move$.
All the clauses are interpreted as follows. $pred1(X,Y)$: $X$ is a block and $Y$ is the top block in a column, where no meaningful interpretation exists; $pred2(X)$: $X$ is a block directly on the floor; 
 $pred4(X,Y)$: $X$ is a block directly on the floor and there is no other blocks above it, and $Y$ is a block; $pred3(X)$: $X$ is the top block in a column that is of at least two blocks in height, which in this task states where the block should be moved to.
The meaning of $move(X,Y)$ is then clear: it moves the movable blocks on the floor to the top of a column that is at least two blocks high.

It is noticeable that the learned policies are sensible but not perfect. A flaw of this policy is that it does not tell what to do when all blocks are on the floor or when there are no movable blocks on the floor, in which case the agent must rely on random moves. In addition, the construction of the policy is not the most concise. The main functionality of $pred4$ is to label the block to be moved, equivalent to the more concise clause: $pred4(X) \leftarrow pred2(X),top(X)$.

\textbf{Induced policies for \textit{Cliff-walking}}: The policies induced by the NLRL agent in the cliff-walking experiment are:
\begin{equation}
    \begin{split}
        &0.990: right() \leftarrow current(X,Y),succ(Z,Y) \\
        &0.561: down() \leftarrow pred(X),last(X) \\
        &0.411: down() \leftarrow current(X,Y),last(X) \\
        &0.988: pred(X) \leftarrow zero(Y),current(X,Z)\\
        &0.653: left() \leftarrow current(X,Y),succ(X,X) \\
        &0.982: up() \leftarrow current(X,Y),zero(Y) \\
    \end{split}
\end{equation}
The agent goes upwards if it is at the bottom row of the whole field.
Actually, the only position the agent needs to move up in the optimal route is the bottom left corner. However, it does not matter here as all other positions in the bottom row are absorbing states.
The agent moves to right if the $Y$ coordinate is larger than 0.
And when the agent reaches the right edge, it will move downwards.
The clause associated to the predicate $left()$ shows that the agent has learned not to move left since there is not a number whose successor is itself. 
There are many other definitions with lower confidence which basically will never be activated.  

Such a policy is a sub-optimal one since it has the chance to bump into the right wall of the field. Though such a flaw is not serious in the training environment, shifting the initial position of the agent to the top left or top right makes it deviate from the optimal policy obviously.
Also, the clause of $down$ can be simplified as $down()\leftarrow current(X,Y),last(X)$, which means move down if the current position is in the rightmost edge.
    \section{Conclusion and Future Work}
In this paper, we propose a novel reinforcement learning method named Neural Logic Reinforcement Learning (NLRL) that is compatible with policy gradient algorithms in deep reinforcement learning. Empirical evaluations show NLRL can learn near-optimal policies in training environments while having superior interpretability and generalizability. In the future work, we will investigate knowledge transfer in the NLRL framework that may be helpful when the optimal policy is quite complex and cannot be learned in one shot. Another direction is to use a hybrid architecture of DILP and neural networks, i.e., to replace $p_S$ with neural networks so that the agent can make decisions using raw sensory data.


\section*{Acknowledgements}
The authors would like to thank Dr Tim Rocktäschel, Dr Frans A. Oliehoek and Gregory Palmer for the helpful discussions, the reviewers for the insightful comments, and Neng Zhang for the proofreading. This work was supported by the EPSRC project ``Robotics and Artificial Intelligence for Nuclear (RAIN)'' (EP/R026084/1).
    \bibliographystyle{icml2019}
    \bibliography{ref}
\end{document}


\section{Differences between DRLM and $\partial$ILP}
There are 2 major differences between DRLM and $\partial$ILP.
First, each deduction step of $\partial$ILP outputs the probabilistic sum of the valuation generated in this step and the valuation in last step, which makes definitions that require less steps of deduction repeatedly affect final outputs, namely, the result valuation.
In RL tasks, this give the agent strong incentive to use simpler strategies with less logic chaining and therefore easier be trapped in local optima.
We instead make the agent treat valuation produced by last step deduction only as input.
The output of each step is simply the sum of deduced valuation and the initial inputs $\boldsymbol{e}_0$ that encloses both the state and background information in NLRL.
Notably, the valuations deduced in last step will not be forgot because they will be derived again in current step.

Another difference is about the weights, in $\partial$ILP, the weights are assigned to combinations of clauses for each intensional predicate, whereas DRLM assign them to individual clauses.
It is easy to see that later approach will use much less variables.
Suppose there are $n$ predicates each defined by $m$ clauses, and each clause has $r$ possible definitions.
The $\partial$ILP implementation will use $nr^m$ variables while the DRLM will only consume $nmr$ of them.
$\partial$ILP used the memory expensive one because they use elementary max to combine valuations of different clauses defining the same predicate, in which case they found conclusions of the two clauses will over-write each other when they are combined and thus the gradient flow is truncated.
To prevent truncated gradient flow, they thus choose to assign variables of combinations of clauses directly.
Such choice makes $\partial$ILP consuming huge amount of memory while at the same time constraint the program templates to define two and only 2 clauses for each intensional predicate.
We use the probabilistic sum in place of max to prevent the truncated gradient flow problem while keeping the advantage of smaller, memory cheaper and more flexible model.

\section{policy interpretation of other tasks}

\textbf{UNSTACK induced policy}: The policy induced by NLRL in \textit{UNSTACK} task is:
\begin{equation}
    \begin{split}
        & 0.972: move(X,Y) \leftarrow isFloor(Y),pred(X) \\
        & 0.987: pred(X) \leftarrow pred2(X),top(X) \\
        & 0.997: pred2(X) \leftarrow on(X,Y),on(Y,Z)
    \end{split}
\end{equation}

We only show the invented predicates that are used by the action predicate and the definition clause with high confidence (larger than 0.3) here.
The $pred2(X)$ means the block $X$ is on top of another block (the block is not directly on the floor).
The $pred(X)$ means the block $X$ is in the top position of a column of blocks and it is not directly on the floor, which basically indicates the block to be moved.
The action predicate $move(X,Y)$ simply move the top block in any column with more than 1 block to the floor.

\textbf{ON induced policy}: The induced policy of the \textit{ON} task is:
\begin{equation}
    \begin{split}
        &1.000: move(X,Y) \leftarrow top(X),pred(X,Y) \\
        &1.000: move(X,Y) \leftarrow top(X),goalOn(X,Y) \\
        &0.947: pred(X,Y) \leftarrow isFloor(Y), pred2(X)\\
        &1.000: pred2(X) \leftarrow on(X,Y),on(Y,Z) \\
    \end{split}
\end{equation}
The goal of \textit{ON} is to move block $a$ onto $b$, while in the training environment the block $a$ is at the bottom of the whole column of blocks.
The strategy NLRL agent learned is to first unstack all the blocks and then move $a$ onto $b$.
The first clause of move $move(X,Y) \leftarrow top(X),pred(X,Y)$ implements the unstack procedures, where the logics are similar to the \textit{UNSTACK} task.
The second clause $move(X,Y) \leftarrow top(X),goalOn(X,Y)$ tells if the block $X$ is already movable (there is no blocks above), just move $X$ on $Y$.
This strategy can deal with most of the circumstances and is optimal in the training environment.
Whereas, we can also construct non-optimal case where unstacking all the blocks are not necessary or if the block $b$ is below the block $a$, e.g., $((b,c,a,d))$.

\textbf{Windy cliff-walking induced policy}: The induced policy of the windy cliff-walking task is:
\begin{equation}
    \begin{split}
        0.999: down()\leftarrow current(X,Y),last(X)
        0.472: right()\leftarrow current(X,Y),succ(Z,Y)
        0.628: right()\leftarrow current(X,Y),succ(Z,X)
        0.966: up()\leftarrow current(X,Y),zero(X)
    \end{split}
\end{equation}
Differing from the original cliff-walking, the agent tends be more conservative.
It tends to move upwards when it is on the left edge of the field.
By doing so, there is less risk of falling to the cliff when it approaching the right edge.